\documentclass[a4,12pt]{article}
\usepackage{latexsym}
\usepackage{amsmath,amstext,amsfonts}
\def\bm#1{\mbox{\boldmath $#1$}}
\def\teigi{\stackrel{\rm def}{=}}

\makeatletter
  
  \@addtoreset{equation}{section}
\makeatother 
\title{Multiplicative Nonholonomic/Newton -like Algorithm  }
\author{Toshinao {\sc
    Akuzawa}\thanks{akuzawa@islab.brain.riken.go.jp}\vspace{0.3cm}\\
and\vspace{0.3cm}\\
Noboru {\sc Murata}
\vspace{0.5cm}\\
Brain Science Institute \\
{\it RIKEN}\\
{\small 2-1 Hirosawa, Wako-shi, Saitama 351-0198, Japan}}
\date{{\it October 19, 1999}}

\begin{document}
\maketitle
\abstract{We construct new algorithms  from scratch, 
which use the  fourth order cumulant  of stochastic 
variables for the cost function.  
The 
multiplicative updating rule 
here constructed  is natural from the
homogeneous nature of the Lie group
and has numerous merits for 
the   rigorous treatment of the 
dynamics.  
As one consequence, the second order convergence is shown.
For the cost function, 
functions invariant under 
the componentwise scaling 
are choosen. 
By
identifying
points  which can be transformed to each other by the scaling, 
 we assume that the dynamics is in a coset space.   
In our method, a point can move toward  any direction in this coset.
Thus, 
 no  prewhitening is  required.   
 }
\section{Introduction}
\label{intro}
Suppose that $N$-dimensional stochastic
variables $\{X_i|1\le i \le N\}$ are observed.
The independent component analysis (ICA) pursues a map
 $X \mapsto Y$, where each component of $Y$ becomes mutually independent.  
In this letter  we restrict ourselves to 
 the linear independent component analysis. 
There  
we want to find a linear transformation $C:{\bf X}=(X_1,\cdots,X_N)'\mapsto
{\bf Y}=(Y_1,\cdots,Y_N)'=C{\bf X}$ which 
 minimizes some cost function that measures the independence. 
Hereafter we  denote by the upper subscript $\prime$ the transposition and
by $\dagger$ the complex conjugate. 

There can be  many candidates for the cost function.  
For example
the Kullback-Leibler information 
is a good measure for the independence. 
In this case 
the problem is translated to 
  the minimization of 
$ -\sum_{i=1}^N\int dy_i P_i(y_i)\ln  P_i(y_i)$, where
$P_i$ is the probability density function of the $i$-th component. 
It is obvious that we must evaluate $P_i$'s to find the optimal
solution. A robust estimation 
of the probability density functions  is not an easy  task 
and if it is possible it may be computationally expensive. 

An alternative idea is to make use of  the cumulant of the fourth
order, or the kurtosis\cite{hyvarinen1}, which we will adopt in this letter. 
The fourth order cumulant vanishes for 
the  normal distribution.  So, this cost function is robust under 
the gaussian random noises. 
We will construct algorithms where a matrix, which specifies the
linear transformation, is updated by the left-multiplication of a 
matrix $D={\rm e}^{\Delta}$. 
This expression implies that $D$ belongs to
$GL(N,{\boldmath R})$ (more accurately, 
the component of $GL(N,{\boldmath R})$ connected to the unit element), 
which ensures the
conservation of the rank.
The specification of $D$ by the coordinate $\Delta$ 
has many advantages 
since it has a compatibility with   the homogeneous nature of the Lie group. 

There are variations for the form of the cost
function. We will show our definitions in the following two sections, which
are choosen to possess  invariance under componentwise scaling. 
This invariance is crucial for 
a rigorous treatment of the convergence properties.  
Moreover, this invariance allows us to 
identify
points  in $GL(N,{\boldmath R})$ which is transformed to each
other by the
scaling.
Then  we can legitimately restrict   the dynamics to a coset space  
which is introduced by this identification.    

Under these settings, we determine $\Delta$ by using the Newton method 
for the second order
expansion of the cost function with respect to $\{\Delta_{ij}\}$. It
is assumed 
that the diagonal elements of $\Delta$ are zeros, 
which does not impose any restrictions. 
That is, a point can move toward  any direction in this coset by a
left-multiplication of ${\rm e}^{\Delta}$. 
Thus 
it is not necesarry for our  method to prewhiten the data. 
It is also not required 
that the
optimal solution is  the maximum or the minimum of the
cost function. Indeed,  the sole requirement is that 
the optimal point  is a saddle point of the cost function 
since our method
is in principle the Newton method.  
These are great advantages of our method. 


Our strategy is as follows.
As an initial condition we set $C_0$. 
For  $t>0~(t\in{\bf N}^{+})$, 
we introduce an  $N\times N$ matrix $\Delta_t$ and 
denote $C_{t}$ as  $C_{t}={\rm e}^{\Delta_{t}}C_{t-1}$.
Next, we evaluate the cost function at $C_{t}$ 
by using the expansion around $C_{t-1}$ 
with respect to the elements of
$\Delta_{t}$ up to the second order. 
Then    $\Delta_t$ is choosen as a saddle point of 
this second order
expansion.  
We iteratively follow these procedures until we obtain a satisfactory
solution.

This letter is organized as follows.
In Section \ref{kurt1} the main part of our algorithm is  constructed, 
where the cost function  is essentially identical to the sum of
kurtoses. 
We adopt the square of the kurtoses for the cost function 
in Section \ref{kurt2}.
Explicit expressions for the optimal 
$\Delta$ (up to the second order)
are obtained both in Sections \ref{kurt1} and \ref{kurt2}. 
 Section \ref{iteration} is a short section  where  we show how
each updating step is combined to obtain the optimal $C$. 
In Section \ref{secconv} the convergence property of our algorithm is 
discussed. Section \ref{disc} contains conclusions and discussions. 
\section{Multiplicative update algorithm}
\label{kurt1}
\subsection{Expansion of the cost function }
Let us start by defining the cost function:
\begin{eqnarray}
  \label{eq:e1}
&&f(C,X)=\sum_i f_i(C,X)~,
  \end{eqnarray}
where $f_i$'s are the fourth order moments 
of components
divided by the square of their variances, 
\begin{eqnarray}
  \label{eq:e1.1}
&&  f_i(C,X)=\frac{E((CX)_i^4)}{E((CX)_i^2)^2}~.
\end{eqnarray}
In this letter we denote by $E(A)$ the expectation  of 
$A$. 
Obviously 
the cost function $f$ coincides with the sum of kurtoses of all the components 
up to  the constant. 
We set $D={\rm e}^{\Delta}$ and
 expand $f(D,Y)$  
 in terms of the elements of $\Delta$. 
For example expansions term  by term are evaluated as follows:
 \begin{eqnarray}
  \label{eq:e2}
E((DY)_i^4)
&=&
E(Y_i^4)+4\sum_{p}(\Delta_{ip}+(\frac{\Delta^2}{2})_{ip})E(Y_i^3Y_p)
+6\sum_{p,q}\Delta_{ip}\Delta_{iq}E(Y_i^2Y_pY_q)+O(\Delta^3)~\nonumber\\
E((DY)_i^2)
&=&
E(Y_i^2)+2\sum_{p}(\Delta_{ip}+(\frac{\Delta^2}{2})_{ip})E(Y_iY_p)
+\sum_{p,q}\Delta_{ip}\Delta_{iq}E(Y_pY_q)+O(\Delta^3)~.
\end{eqnarray}
Hereafter we denote by 
 $O(\Delta^k)$   polynomials of matrix elements of $\Delta$ which 
does not contain terms with degrees less than $k$. 
For  brevity's sake  
we introduce the following notations:
\begin{eqnarray}
  \label{eq:e3.1}
&&  \sigma_i^{(k)}=|E(Y_i^k)|^{1/k}~,\\
&&  R^{(k)}_{pi}=\frac{E(Y_i^k Y_p)}{(\sigma^{(2)}_i)^{k+1}}~,\\
&&  U^{(k,i)}_{pq}=\frac{E(Y_i^kY_p Y_q)}{(\sigma^{(2)}_i)^{k+2}}~,
\end{eqnarray}
and
\begin{eqnarray}
  \label{eq:e3.2}
  && \kappa_i={(\sigma^{(4)}_i)^4}/{(\sigma^{(2)}_i)^4}~.
\end{eqnarray}
Using the quantities defined above we can  show that  the
cost function is expanded as 
\begin{eqnarray}
  \label{eq:e4}
 f_i(D,Y)
&=&\bigg[
\kappa_i+4\big[(\Delta+\frac{\Delta^2}{2})R^{(3)}\big]_{ii}
+6\big[
\Delta U^{(2,i)}\Delta'
\big]_{ii}
+O(\Delta^3)
\bigg]\nonumber\\
&&~~\times
\bigg[
1-4\big[(\Delta+\frac{\Delta^2}{2})R^{(1)}\big]_{ii}
-2\big[
\Delta U^{(0,i)}\Delta'
\big]_{ii}
+12\big[
\Delta R^{(1)}
\big]_{ii}^2
+O(\Delta^3)
\bigg]\nonumber\\
&=&\kappa_i - 4\big[(\Delta+\frac{\Delta^2}{2})(\kappa_i
R^{(1)}-R^{(3)})\big]_{ii}
+2\big[
\Delta (3U^{(2,i)}-\kappa_i  U^{(0,i)})\Delta'
\big]_{ii}\nonumber\\
&&~~
+12\kappa_i\big[
\Delta R^{(1)}
\big]_{ii}^2
-16\big[
\Delta R^{(1)}
\big]_{ii}\big[
\Delta R^{(3)}
\big]_{ii}+O(\Delta^3)~
\end{eqnarray}
by  straightforward calculations. 
Next, we evaluate  partial derivatives of the cost function 
by the matrix elements of $\Delta$. 
 Partially differentiating  (\ref{eq:e4}), 
we get an expression, 
\begin{eqnarray}
  \label{eq:e5}
&&  \frac{\partial f({\rm e}^{\Delta},Y)}{\partial \Delta_{kl}}=
-4\big[K-R^{(3)}\big]_{lk}
-2\big[(K-R^{(3)})\Delta+\Delta(K-R^{(3)})\big]_{lk}\nonumber\\
&&+4\big[
 (3U^{(2,k)}-\kappa_k  U^{(0,k)})\Delta'
\big]_{lk}
+24K_{lk}\big[\Delta R^{(1)}
\big]_{kk}
-16R^{(1)}_{lk}\big[\Delta R^{(3)}
\big]_{kk}
-16 R^{(3)}_{lk}\big[\Delta R^{(1)}
\big]_{kk}\nonumber\\
&&+O(\Delta^2)~,
\end{eqnarray}
where $K$ is an $N\times N$ matrix defined by
\begin{eqnarray}
  \label{eq:e5.9}
&&K_{pq}=\kappa_q  R^{(1)}_{pq}~.  
\end{eqnarray}
We want to decide $\Delta$  for which 
 the partial derivative 
by  $\Delta_{kl}~(k\ne
 l)$
of the cost function 
 vanish on condition that 
 $\Delta_{ii}=0$ for $1\le i \le N$.  
We neglect $O(\Delta^3)$ terms in the cost function. 
Thus the  right-hand side of (\ref{eq:e5}) is 
regarded as a polynomial of
 $\{\Delta_{kl}\}$ 
of at most first order and it is  always possible 
in principle to
 determine $\Delta$ which satifies the above condition. 
It is, at the same time, not easy  to  describe  the problem 
in a  form which is valid 
for
arbitrary $N$. 
In the following subsection we will introduce  a transparent and unified 
method for handling the partial derivatives of $f$. 
We leave this subsection by
introducing $N\times N$ matrices 
\begin{eqnarray}
  \label{eq:e6}
&&  V^{(i)}=3U^{(2,i)}-\kappa_i  U^{(0,i)}~
\end{eqnarray}
and
\begin{eqnarray}
  \label{eq:e6.1}
 &&  Q=K-R^{(3)}~
\end{eqnarray}
for later convenience.  
\subsection{Expression by tensor product and determination of $\Delta$}
The expression (\ref{eq:e5}) is quite  complicated and   not
convenient for our purpose, 
`` determine $\Delta$, where
all the partial derivatives  vanish''. 
Fortunately by  mapping  the relations between elements of 
$N\times N$ matrices  to those of   $N^2\times
N^2$ matrices, we can handle the problem transparently. 
Some preparations
are needed.
First, let us introduce a map $\rm cs$:
\begin{eqnarray}
  \label{eq:a14}
  {\rm Mat}(N,{\boldmath F}) &\rightarrow& {\boldmath F}^{N^2}\nonumber\\
A=\left(
  \begin{array}{cccc}
 A_{11}& A_{12}&\cdots &A_{1N}\\
A_{21} &\multicolumn{3}{c}{\dotfill}\\
\multicolumn{4}{c}{\dotfill}\\
A_{N1} &\multicolumn{2}{c}{\dotfill}&A_{NN}
  \end{array}
\right) &\mapsto& 
{\rm cs}(A)=
(A_{11}~ A_{21}~ \cdots~ A_{N1}~ A_{12}~ A_{22}~\cdots~ A_{NN})'~,\nonumber\\
\end{eqnarray}
where $\boldmath F$ is an unspecified  field. 
We also introduce 
two useful operators $T$ and $P$. 
The ``intertwiner'' $T$ is  an $N^2\times N^2$ matrix 
defined by 
\begin{eqnarray}
  \label{eq:a15}
  {\rm cs}(A')=T{\rm cs}(A) ~\mbox{\rm for~} A\in  {\rm Mat}(N,{\boldmath F})~.
\end{eqnarray}
The projection  operator $P$,
\begin{eqnarray}
  \label{eq:a18}
P&=&{\rm diag}(p_1,\cdots,p_{N^2})~,\nonumber\\
&&\left\{
\begin{array}{ll}
 p_k=1 ~~~\mbox{\rm for}~~ k=N(i-1)+i,1\le i\le N~\\
 p_k=0~~~~ \mbox{\rm otherwise}~,
\end{array}
\right.
\end{eqnarray}
 is used to extract the ``diagonal''
elements of a matrix from its image by $\rm cs$. 

On this setting we can rewrite (\ref{eq:e5}) as
\begin{eqnarray}
  \label{eq:e7}
  \frac{\partial f({\rm e}^{\Delta},Y)}{\partial \Delta_{kl}}&=&
\bigg[ -4{\rm cs}(Q)
-2\big[I_N\otimes Q+T(I_N\otimes Q')T\big]{\rm cs}(\Delta)
+4
\big\{\bigoplus_{i=1}^N V^{(i)}\big\}
{\rm cs}(\Delta')
\nonumber\\&&
+
\bigg\{24(I_N \otimes K)P(I\otimes R^{(1)})'
-16 ( I_N \otimes R^{(1)})P(I\otimes R^{(3)})'\nonumber\\
&&-16 (I_N\otimes R^{(3)})P(I\otimes R^{(1)})'
\bigg\}
{\rm cs}(\Delta')
\bigg]_{l+N(k-1)}~,
\end{eqnarray}
where $I_N$ is the $N\times N$ unit matrix and
\begin{eqnarray}
  \label{eq:tiu1}
\bigoplus_{i=1}^N V^{(i)}=
\left(
  \begin{array}{lllll}
V^{(1)} & 0 & \multicolumn{2}{c}{\cdots\cdots} & 0\\ 
0& V^{(2)} & 0 & \multicolumn{2}{c}{\cdots\cdots}\\
 \multicolumn{5}{c}{\dotfill}\\
 \multicolumn{5}{c}{\dotfill}\\
 0& \multicolumn{2}{c}{\cdots\cdots}& V^{(N-1)}& 0   \\
0& 0& \multicolumn{2}{c}{\cdots\cdots}& V^{(N)}   \\
   \end{array}
\right)~.
\end{eqnarray}
We make use of the following fact:\\
For $X\in {\rm Mat}(N,{\boldmath F})$
\begin{eqnarray}
  \label{eq:e8f}
  T(I_N\otimes X)T=X\otimes I_N~.
\end{eqnarray}
See  Appendix \ref{app:prf} for the proof of  (\ref{eq:e8f}). 
Then (\ref{eq:e7}) becomes 
\begin{eqnarray}
  \label{eq:e77}
&&  \frac{\partial f({\rm e}^{\Delta},Y)}{\partial \Delta_{kl}}=
 -4[{\rm cs}(Q)]_{l+N(k-1)}
+\big[
W
{\rm cs}(\Delta)
\big]_{l+N(k-1)}~,\nonumber\\
\end{eqnarray}
where
\begin{eqnarray}
  \label{eq:e8}
W&=&
-2\big(I_N\otimes Q+Q'\otimes I_N\big)
+4
\big\{\bigoplus_{i=1}^N V^{(i)}\big\}
T
+
\bigg[24(I_N\otimes K)P(I\otimes R^{(1)})'
\nonumber\\&&
-16 (I_N \otimes R^{(1)})P(I\otimes R^{(3)})'
-16 (I_N \otimes R^{(3)})P(I\otimes R^{(1)})'
\bigg]
T~.\nonumber\\
\end{eqnarray}
Now let us determine $\Delta$. 
Remember that we are going along the spirit of the Newton method. 
Thus we want to find $\Delta$ which satisfies
the  conditions
\begin{eqnarray}
  \label{eq:e10}
    \frac{\partial f({\rm e}^{\Delta},Y)}{\partial
    \Delta_{kl}}=0+O(\Delta^2)~~
\mbox{\rm for } 1\le k,l \le N,~k\ne l
\end{eqnarray}
and 
\begin{eqnarray}
  \label{eq:e11}
  \Delta_{kk}=0 ~~\mbox{\rm for}~~ 1\le k\le N~.
\end{eqnarray}
The conditions (\ref{eq:e11}) make the problem rather complicated one.
Fortunately, 
by using $P$ 
we can combine 
the conditions  (\ref{eq:e10}) and  (\ref{eq:e11}) into 
  a matrix equation :
\begin{eqnarray}
  \label{eq:e19}
\Big[(I_{N^2}-P)
W(I_{N^2}-P)
+P
\Big]
{\rm cs}(\Delta)-4(I_{N^2}-P){\rm cs}(Q)=0~.
\end{eqnarray}
Immediately it follows that  
\begin{eqnarray}
  \label{eq:e20}
{\rm cs}(\Delta)=4
\Big[(I_{N^2}-P)
W
(I_{N^2}-P)
+P
\Big]^{-1}
(I_{N^2}-P){\rm cs}(Q)~.
\end{eqnarray}
Thus we have obtained $\Delta$ which specify a saddle point of 
the  expansion of 
$f(C,Y)$ up to the second order.
Note that quantities in the right-hand side of (\ref{eq:e20}) are easily estimated
ones 
from the
observed data.
So, an updating is determined by (\ref{eq:e20}) without any
ambiguities. 

\section{Case $\rm I\!I$:  square of kurtosis}
\label{kurt2}
Obviously, points where kurtosis 
vanishes do not play any special role  for
the cost function  $f$ in Section \ref{kurt1}. The optimal solution, however, 
contains components with zero kurtoses 
when the number of the sources is less than that of the observation channels. 
Thus, 
in this section  we treat with  a slightly different
  cost function, which  is the sum,
\begin{eqnarray}
  \label{eq:se1}
&&{\bm f}(C,X)=\sum_i {\bm f}_i(C,X)~,
  \end{eqnarray}
of the square of the kurtoses, 
\begin{eqnarray}
  \label{eq:se1.1}
&&  {\bm f}_i(C,X)=\left[\frac{E((CX)_i^4)}{E((CX)_i^2)^2}-3\right]^2~.
\end{eqnarray}
As in the last section, we want to know the saddle point 
$D={\rm  e}^{\Delta}$ of 
the expansion of ${\bm
  f_i}(D,Y)$ in 
terms of $\{\Delta_{ij}\}$ up to the second order.
We do not describe details of the calculations in this section, 
which is 
 carried out 
almost in the same way as in Section \ref{kurt1}. 
First, the expansion of ${\bm
  f_i}(D,Y)$ is evaluated as 
\begin{eqnarray}
  \label{eq:se4}
 {\bm f}_i(D,Y)
&=&(\kappa_i-3)^2 - 8\big[(\Delta+\frac{\Delta^2}{2})(
R^{(1)}\kappa_i-R^{(3)})\big]_{ii}(\kappa_i-3)\nonumber\\
&&+4\big[
\Delta (3U^{(2,i)}-\kappa_i  U^{(0,i)})\Delta'
\big]_{ii}(\kappa_i-3)
+16\big[
\Delta (R^{(1)}\kappa_i-R^{(3)})
\big]_{ii}^2
\nonumber\\
&&
+24(\kappa_i-3)\kappa_i\big[
\Delta R^{(1)}
\big]_{ii}^2
-32(\kappa_i-3)\big[
\Delta R^{(1)}
\big]_{ii}\big[
\Delta R^{(3)}
\big]_{ii}+O(\Delta^3)~.
\end{eqnarray}
Next, we introduce  $N\times N$ matrices $\bm K$, $\{{\bm
  V}^{(i)}|1\le i\le N\}$, 
$\bm S$, and $\bm Q$ 
defined respectively by
\begin{eqnarray}
  \label{eq:se5.9}
&&{\bm K}_{pq}=  2R^{(1)}_{pq}(\kappa_q-3)\kappa_q~,  
\end{eqnarray}
\begin{eqnarray}
  \label{eq:se6}
&&  {\bm V}^{(i)}=2(\kappa_i-3)(3U^{(2,i)}-\kappa_i  U^{(0,i)})~,\\
\end{eqnarray}
\begin{eqnarray}
  \label{eq:se6.001}
  {\bm S}={\rm diag}(2(\kappa_i-3))~,
\end{eqnarray}
and
\begin{eqnarray}
  \label{eq:se6.1}
 && {\bm Q}_{pq}=2(\kappa_q-3)(R^{(1)}_{pq}\kappa_q-R^{(3)}_{pq})~.
\end{eqnarray}
We also rewrite $Q$ in (\ref{eq:e6.1}) by $\bm q$ in order to avoid confusions:
\begin{eqnarray}
  \label{eq:se6.2}
 && {\bm q}_{pq}=(R^{(1)}_{pq}\kappa_q-R^{(3)}_{pq})~.
\end{eqnarray}
Now 
we proceed to the expression by using the tensor product.
We can show  that the gradients of the cost function have the
following expression: 
\begin{eqnarray}
  \label{eq:se77}
&&  \frac{\partial {\bm f}({\rm e}^{\Delta},Y)}{\partial \Delta_{kl}}=
 -4[{\rm cs}({\bm Q})]_{l+N(k-1)}
+\big[
{\bm W}
{\rm cs}(\Delta)
\big]_{l+N(k-1)}+O(\Delta^2)~,\nonumber\\
\end{eqnarray}
where
\begin{eqnarray}
  \label{eq:se8}
{\bm W}&=&
-2\big(I_N\otimes {\bm Q}+{\bm Q'}\otimes I_N\big)
+4
\big\{\bigoplus_{i=1}^N {\bm V}^{(i)}\big\}
T
+
\bigg[24( I_N\otimes {\bm K})P(I\otimes R^{(1)})'
\nonumber\\&&
+32( I_N\otimes {\bm q})P(I_N\otimes {\bm q})'
-16 ( I_N\otimes R^{(1)}{\bm S})P(I\otimes R^{(3)})'
\nonumber\\&&
-16 ( I_N\otimes R^{(3)}{\bm S})P(I\otimes R^{(1)})'
\bigg]
T~.
\end{eqnarray}
This is a  completely analogous expression  to  (\ref{eq:e77}).  
Thus  the coordinate $\Delta$ of the saddle point of the second order 
expansion  
is determined by
\begin{eqnarray}
  \label{eq:se20}
{\rm cs}(\Delta)=4
\Big[(I_{N^2}-P)
{\bm W}
(I_{N^2}-P)
+P
\Big]^{-1}
(I_{N^2}-P){\rm cs}({\bm Q})~.
\end{eqnarray}
In many cases obtained through the two cost functions in Section
\ref{kurt1} and Section \ref{kurt2}  are almost the same results.  
As  implied at the beginning of this section, 
the main difference between these two lies in the points where the kurtosis of
one of the components vanishes. 
These point indeed constitue saddle points of 
 the   cost function
$\boldmath f$, while  it is impossible to  capture them by the
algorithm in Section \ref{kurt1}. 
Thus, we must choose an appropriate method for individual problems 
having this differnce in mind. 

\section{Iteration of updating}
\label{iteration}
Now we have obtained the updating rules. It is not necessary to tune the
learning rate. Apparently, (\ref{eq:e19}) 
and (\ref{eq:se20})
look complicated. 
They are, however, easily implemented by the numerical tools like MatLab. 
(The source codes will be available from our Web-site. )
Starting from $C_0$, 
$C_i$ for positive $i$ is determined by the left multiplication by 
${\rm e}^{\Delta_i}$, where 
$\Delta$ is determined by setting $Y=C_{i-1}X$,  
i.e,
\begin{eqnarray}
  \label{eq:b1}
  C_t={\rm e}^{\Delta_{t}}{\rm e}^{\Delta_{t-1}}{\rm e}^{\Delta_{t-2}}\cdots{\rm e}^{\Delta_{1}}C_0~.
\end{eqnarray}
If $\Delta$ becomes saficiently small, we can stop the iteration and exit the 
process. 

\section{Second order convergence}
\label{secconv}
First, we will take over  the notations in Section \ref{kurt1}. 
The following discussion  is, however, valid for the algorithm in Section
\ref{kurt2} if we  substitute the quantities  $f$,  $W$, and so on by 
their boldface counterparts. 
Let us  start this section by introducing some additional notations. 
We set 
\begin{eqnarray}
  \label{eq:pr1}
  G\in GL(N,{\boldmath R})
\end{eqnarray}
and 
\begin{eqnarray}
  \label{eq:prd2}
  K\in GL(1,{\boldmath R})^{\oplus N}~.
\end{eqnarray}
We also define the coset space  $K\backslash G$ by
introducing  the equivalence relation 
\begin{eqnarray}
  \label{eq:pr3}
g' g^{-1}\in K
\Longleftrightarrow 
 g\sim g'
\end{eqnarray}
to $G$. That is, $K\backslash G\cong\{Kg|g\in G\}$. 
Our method is 
understood as 
an orthodox adaptation of the Newton method to this 
coset space $K\backslash G$. 
Note that  
the cost function $F(\cdot)\teigi f(\cdot,Y)$ on $G$ 
 satisfies the relation 
\begin{eqnarray}
  \label{eq:pr4}
F(g)=F(Kg)~.
\end{eqnarray}
So $F$ is  naturally considered as a function on $K\backslash G$. 
That is the reason of our choice for  the cost function. 
Thus, the second-order convergence immediately follows if the 
the correction to the  error with respect to the  coordinating
resulting from the  multiplicative nature is properly evaluated.

At time $t$, a point $g$ on $K\backslash G$ is specified by
the coordinate $X^{(t)}(g) \in{\frak m}$ such that 
\begin{eqnarray}
  \label{eq:prf101}
  {\rm e}^{X^{(t)}(g)}C_t\sim g~,
\end{eqnarray}
where $\frak m$ is the set of $N\times N$ matrices whose diagonal
elements are zeros. 
Actually, this statement itself  is not a thing of course, for which the proof
will be given
elsewhere. 
Define $F_t$, the representation of the cost function at $t$,   by
\begin{eqnarray}
  \label{eq:prf102}
  F_t(X)=F(  {\rm e}^{X}C_t)~.
\end{eqnarray}
Here we introduce an $(N^2-N)\times N^2$ matrix $\tilde P$ by
drawing out the $i+N(i-1)$-th raws from the unit $N^2\times N^2$
matrix where $i=N,N-1,\cdots, 2,1$. 
We will denote by $\boldmath H^{(t)}$ the  Hessian, 
\begin{eqnarray}
  \label{eq:prf102.11}
  {\boldmath H}^{(t)}_{kl}=\frac{\partial^2 F_t(X)}
{\partial ({\tilde P}{\rm cs}(X))_k\partial ({\tilde P}{\rm cs}(X))_l}
\end{eqnarray}
Note that if we set
\begin{eqnarray}
  \label{eq:prf103}
h_t(X)=\left.
T\bigg((I_{N^2}-P)
W(I_{N^2}-P)
+P\bigg)\right|_{C={\rm e}^X C_t}~,
\end{eqnarray}
the Hessian is written as
\begin{eqnarray}
  \label{eq:prf103.1}
  {\boldmath H}^{(t)}={\tilde P}h_t{\tilde P}' ~. 
\end{eqnarray}
Suppose that at some neighborhood of the optimal solution $g_*$,
${\boldmath H}^{(t)}(X)$ 
is Lipschitz continuous for some $t$:
\begin{eqnarray}
  \label{eq:prf104}
  ||{\boldmath H}^{(t)}(X)-{\boldmath H}^{(t)}(X')||\le L ||X-X'||~,
\end{eqnarray}
where $||A||$ is the norm of a  matrix $A$ as the Euclidian space,
\begin{eqnarray}
  \label{eq:norm1}
  ||A||^2={\rm tr}(AA^{\dagger})~.
\end{eqnarray}
We set
\begin{eqnarray}
  \label{eq:prf104.001}
\beta=||H^{(t)}(X^t(g_*))^{-1} ||  ~.
\end{eqnarray}
There exists a positive real number $r$, 
 for which 
\begin{eqnarray}
  \label{eq:prf104.002}
  ||H^{(t)}(X^t(g))^{-1} || <2\beta~~\mbox{\rm for}~
\forall g\in B^{(t)}(g_*,r)\teigi\bigg\{g\bigg|r> ||X^t(g)-X^t(g_*)||~\bigg\}
\end{eqnarray}
 is satisfied. 
Then 
it is known that 
for all $g\in B(g_*,{\rm min}(r,(2\beta L)^{-1}))$, 
\begin{eqnarray}
  \label{eq:prf104.003}
  ||X^t(C_{t+1})-X^t(g_*)||\le  \beta L ||X^t(C_{t})-X^t(g_*)||^2
\end{eqnarray}
and
\begin{eqnarray}
  \label{eq:prf104.004}
  ||X^t(C_{t+1})-X^t(g_*)||\le \frac{1}{2} ||X^t(C_{t})-X^t(g_*)||
\end{eqnarray}
are fulfilled. Thus the second order convergence in this norm  is shown.
Unfortunately, this norm is not invariant and is  unnatural. 
(A natural   metric on $K\backslash G$
 is  one which is  invariant  under the parallel transformation,  
which is induced by the action
 of  elements in $K\backslash G$
 from the right-hand side.) But, it suffices in practice.

\section{Discussions}
\label{disc}
\subsection{Nonholonomy?}
Our method is  related to the nonholonomic method 
by 
Amari, Chen, and Chichocki\cite{amari-chen-cichocki1}. 
In essence our method is a Newton
 approach to the same problem, the optimization without prewhitening. 
Let us   set 
\begin{eqnarray}
  \label{eq:conc11}
  {\rm e}^{z} = {\rm e}^{x}{\rm e}^{y}
\end{eqnarray}
for $x,y\in {\frak gl}(N,{\boldmath R})$.  
Then it is obvious that $z$ does not necessarily belongs to $\frak m$ 
 even if $x,y\in {\frak m}$(, that is,
 $z_{ii}$'s do not always  vanish
when $x_{ii}=y_{ii}=0$ for $1\le i\le N$).  
This may be explained by using the concept of nonholonomy. 
 The degree of freedom in each step, however, equals the dimension
of the space $K\backslash G$ in our setting. The nonholonomic nature 
emerges when we go back to $G=GL(N,{\bm R})$ again. 

There exist several
studies\cite{takeuchi1,helgason1,helgason2,helgason3,akuzawa5} which
deal with
cosets
like $K\backslash G$  or the right coset $G/K$
 when $K$ is a maximal compact subgroup of $G$. 
Unfortunately, what we are studying is the case where $K$ is not a
maximal compact subgroup of $G$. 
So, for example
it is necessary to show 
 whether the  coordinate (\ref{eq:prf101})  is justified or not. 
As mentioned above, further studies including this justification 
 will  appear  elsewhere. 

\subsection{Global convergence}
We should carefully treat  
 first few  steps since this method also has 
a somewhat undesirable global convergent property  inherent in 
the  Newton method. Fortunately enough, 
there exist methods which can 
handle the earlier stage. For example, the nonholonomic gradient
 method\cite{amari-chen-cichocki1}  
may be applicable. 
Another posiibility is to construct a nonholonomic fixed-point
 algorithm which uses the kernel method. 
These methods are suitable for  capturing the optimal point which
 contains components with zero kurtoses. There
 we must, of course,     use the method in Section \ref{kurt2}.  
If it is not necessary to worry about these zero kurtosis components, 
there is little difference between the two methods described in 
Section \ref{kurt1} and Section \ref{kurt2}. 

\subsection{Conclusions}
We have constructed a new  algorithm for finding a optimal point in a
matrix space, where we have  introduced a new   
multiplicative updating method. 
%
The algorithm is in essence the Newton method on a
coset. 
So it converges quite rapidly and it can capture the saddle point. 
Since it does not require prewhitening, 
it is not necessary to worry about the error resulting from the 
prewhitening.  
Indeed, it is  possible to increase
the kurtosis slightly  for data preprocessed by 
the FastICA\cite{fastica1}.

\appendix
\section*{appendix}
\section{proof  of (\ref{eq:e8f})}
\label{app:prf}
\begin{quote}
For $B\in GL(N,{\boldmath F})$ and $1\le i,j\le N$,
\begin{eqnarray}
  \label{eq:proof1}
[  T(X\otimes Y)T {\rm cs}(B)]_{i+N(j-1)}
&=&[  (X\otimes Y)T {\rm cs}(B)]_{j+N(i-1)}\nonumber\\
&&=X_{ip}Y_{jq}(B')_{qp}
=(YB'X')_{ji}~.
\end{eqnarray}
On the other hand 
\begin{eqnarray}
  \label{eq:proof12}
[  (Y\otimes X) {\rm cs}(B)]_{i+N(j-1)}
&&=Y_{jp}X_{iq}B_{qp}
=(YB'X')_{ji}~.
\end{eqnarray}
This proves the statement since $\rm cs$  is bijective. $\Box$
\end{quote}


\begin{thebibliography}{8}

\bibitem[A.Hyv\"arinen,1997]{hyvarinen1}
A.Hyv\"arinen (1997).
\newblock A Fast Fixed-Point Algorithm for Independent Component Analysis.
\newblock {\em Neural Computation\/}, {\em 9\/}, 1483--1492.

\bibitem[Amari {\em et~al.\/},1997]{amari-chen-cichocki1}
Amari, S., Chen, T.-P., \& Cichocki, A. (1997).
\newblock Non-holonomic Constraints in Learning Algorithms for Blind Source
  Separation.
\newblock {\em preprint\/}.

\bibitem[Hurri {\em et~al.\/},1998]{fastica1}
Hurri, J., G\"avert, H., S\"alel\"a, J., \& Hyv\"arinen, A. (1998).
\newblock FastICA package for MATLAB.
\newblock http://www.cis.hut.fi/projects/ica/fastica/.

\bibitem[M.Takeuchi,1994]{takeuchi1}
M.Takeuchi (1994).
\newblock {\em Modern Spherical Functions\/}.
\newblock Amer. Math. Soc.

\bibitem[S.Helgason,1962]{helgason2}
S.Helgason (1962).
\newblock {\em Differential Geometry and Symmetric Spaces\/}.
\newblock Academic Press.

\bibitem[S.Helgason,1978]{helgason1}
S.Helgason (1978).
\newblock {\em Differential Geometry, Lie Groups and Symmetric Spaces\/}.
\newblock New York: Academic Press.

\bibitem[S.Helgason,1984]{helgason3}
S.Helgason (1984).
\newblock {\em Groups and Geometric Analysis\/}.
\newblock Academic Press.

\bibitem[T.Akuzawa \& M.Wadati,1998]{akuzawa5}
T.Akuzawa \& M.Wadati (1998).
\newblock Diffusions on symmetric spaces of type A${\rm I\!I\!I}$ and random
  matrix theories for rectangular matrices.
\newblock {\em J.Phys.A\/}, {\em 31\/}, 1713--1732.

\end{thebibliography}
\end{document}